\documentclass[runningheads]{llncs}
\usepackage[T1]{fontenc}
\usepackage{graphicx}
\usepackage{cite}
\usepackage{amsmath,amssymb,amsfonts}
\usepackage{algorithmic}
\usepackage{textcomp}
\usepackage{xcolor}
\usepackage{fancyhdr}
\usepackage{mathcomp}
\usepackage{siunitx}
\usepackage{algorithm}
\usepackage{stfloats}

\usepackage{wrapfig}

\begin{document}
\title{Single and Multi-Objective Optimization Benchmark Problems Focusing on Human-Powered Aircraft Design}
%
%
\author{Nobuo Namura\orcidID{0000-0002-3523-0389}}
\authorrunning{N. Namura}
%
\institute{Fujitsu Limited, Kawasaki 211-8588, Japan\\
\email{namura.nobuo@fujitsu.com}}
\maketitle              
\begin{abstract}
The landscapes of real-world optimization problems can vary strongly depending on the application. In engineering design optimization, objective functions and constraints are often derived from governing equations, resulting in moderate multimodality. However, benchmark problems with such moderate multimodality are typically confined to low-dimensional cases, making it challenging to conduct meaningful comparisons. To address this, we present a benchmark test suite focused on the design of human-powered aircraft for single and multi-objective optimization. This test suite incorporates governing equations from aerodynamics and material mechanics, providing a realistic testing environment. It includes 60 problems across three difficulty levels, with a wing segmentation parameter to scale complexity and dimensionality. Both constrained and unconstrained versions are provided, with penalty methods applied to the unconstrained version. The test suite is computationally inexpensive while retaining key characteristics of engineering problems. Numerical experiments indicate the presence of moderate multimodality, and multi-objective problems exhibit diverse Pareto front shapes.
\keywords{benchmark \and multi-objective optimization \and aircraft design.}

\end{abstract}
%
%
%

\section{Introduction}
Various methods for black-box optimization, such as evolutionary algorithms (EAs) and Bayesian optimization (BO), have been widely applied across fields, including engineering design, facility operation, material development, drug discovery, and machine learning. The landscapes of objective functions in these applications have become increasingly diverse. In engineering design optimization, objective functions often exhibit moderate multimodality due to governing equations, such as partial differential equations, which typically result in smooth landscapes \cite{kanazaki2013wind, chernukhin2013multimodality, bons2019multimodality, khurana2009airfoil}. The successful use of BO with Gaussian process (GP) surrogate models in this field can be attributed to these characteristics \cite{sakata2003structural, zuhal2019comparative}.

Conversely, the development and evaluation of algorithms for black-box optimization commonly rely on synthetic benchmark problems, which are defined by mathematical equations. These benchmarks significantly influence the research direction of optimization algorithms. In single-objective optimization, fundamental functions like the sphere and Rastrigin function, as well as composite functions formed from these, are frequently used \cite{li2013benchmark}. In multi-objective optimization, benchmark problems often rely on similar fundamental functions \cite{zdt, dtlz}, as they impact convergence toward the Pareto front (PF). 
These fundamental functions can generally be classified into two categories: those with simple unimodal landscapes (e.g., sphere, ellipsoid) and those with complex landscapes that exhibit an exponential increase in local optima as the dimensionality grows (e.g., Rastrigin, Styblinski-Tang). Benchmark problems with moderate multimodality (e.g., Branin, Hartmann), are typically available only in lower dimensions. 
Recent studies \cite{yazdani2023gnbg, schapermeier2023peak} have proposed adjustable benchmark problems  which create multimodality by combining multiple simple basis functions like ellipsoid functions. However, their landscape can be oversimplified due to the small number of basis functions to achieve moderate multimodality. 

In multi-objective cases, benchmark problems possess characteristics that differ from real-world problems, raising concerns about overfitting algorithms to these benchmarks \cite{ishibuchi2016performance}. Ishibuchi et al. \cite{ishibuchi2019} noted that many synthetic benchmark problems lack unique optimal solutions for each objective, and in three-objective problems, the PF often forms a triangle. To address this, benchmark problems like Minus-DTLZ and Minus-WFG \cite{ishibuchi2016performance}, which generate PFs resembling the inverted triangles frequently observed in real problems, have gained popularity.

In recent years, practical benchmark problems based on real-world problems \cite{tanabe2020easy, wang2018batched, zapotecas2023engineering} have been used instead of synthetic benchmark problems. While these problems exhibit complex PF shapes, making it challenging to obtain diverse Pareto-optimal solutions, they are typically available only in low-dimensional spaces up to 10 dimensions. This has raised concerns that convergence to the PF may be too easy \cite{pang2024analysis}.
Additionally, they often require computationally expensive numerical simulations \cite{daniels2018suite, volz2019single, he2020repository}, which can limit the number of function evaluations. 
Therefore, benchmark problems with moderate multimodality that mimic engineering design optimization and allow dimensionality changes while keeping evaluation costs low are scarce. Setting up appropriate benchmark problems under these conditions can be challenging. 

In this paper, we propose single and multi-objective benchmark problems related to the aerostructural design of human-powered aircraft (HPA), which emulate engineering design optimization tasks. The objective functions and constraints are derived from integral equations in aerodynamics and differential equations in material mechanics. These benchmark problems are designed to reflect the characteristics of real-world engineering design optimization, while ensuring short evaluation times. Additionally, by adjusting the definition of design variables, we introduce three levels of difficulty and enable changes in dimensionality through parameter modifications, making these problems scalable. While the original benchmark problems include constraints, we also provide unconstrained versions using penalty methods. Python source code for these benchmark problems is available on GitHub\footnote{https://github.com/Nobuo-Namura/hpa}.

\section{Benchmark Problem Definition}

\subsection{Definition of Variables for HPA Design}
\begin{figure}[t]
\centering
\begin{minipage}[b]{0.49\columnwidth}
    \centering
    \includegraphics[width=0.95\columnwidth]{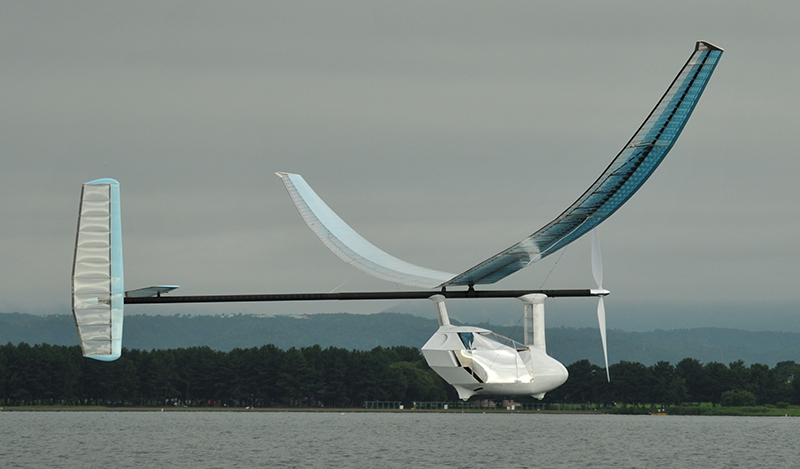}
    \caption{Human-powered aircraft.}
    \label{fig::hpa}
\end{minipage}
\begin{minipage}[b]{0.49\columnwidth}
    \centering
    \includegraphics[width=0.95\columnwidth]{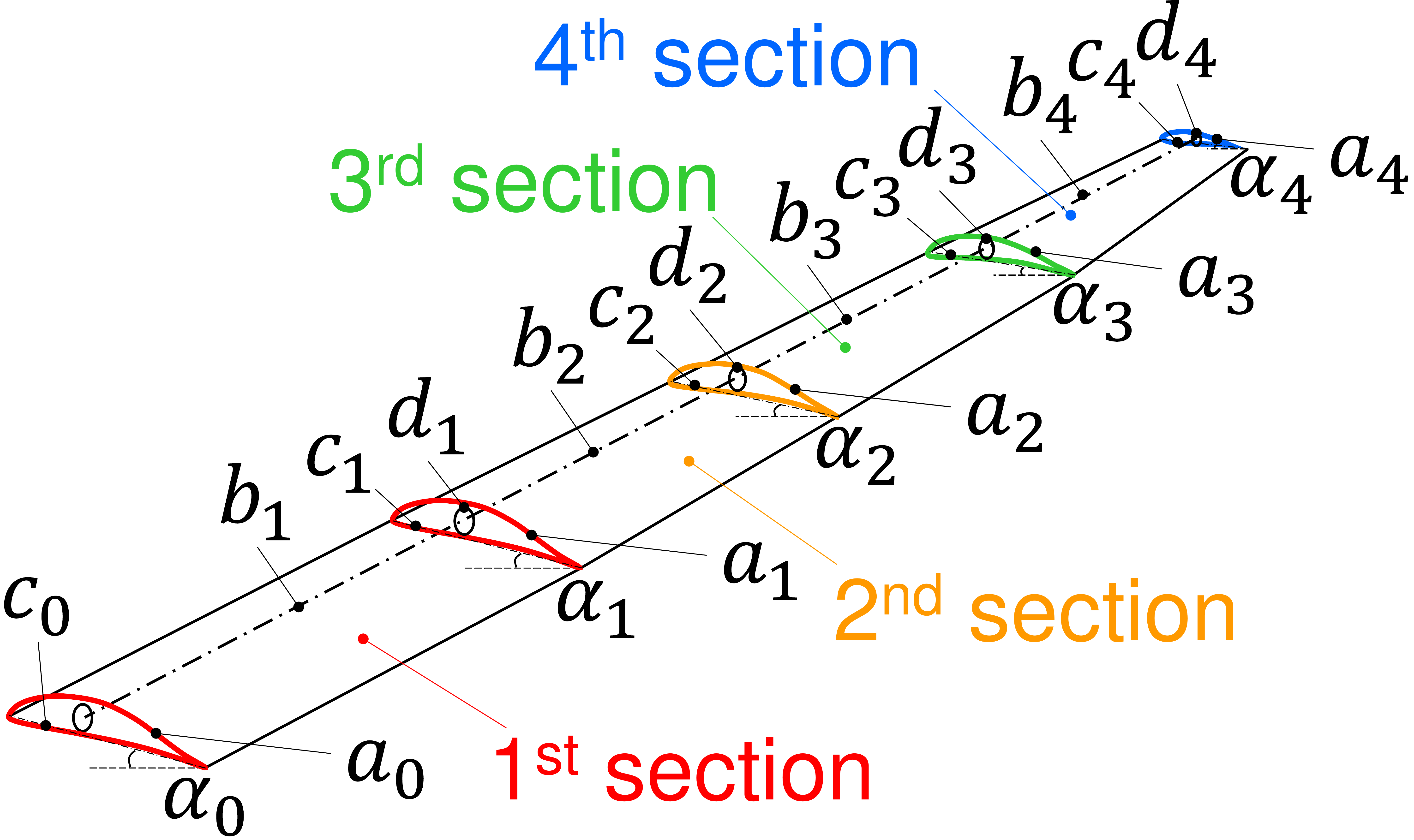}
    \caption{Wing shape definition.}
    \label{fig::variables}
\end{minipage}
\end{figure}

An HPA gains propulsion by the pilot pedaling like a bicycle to rotate a propeller, as depicted in Fig. \ref{fig::hpa}. 
Wings of HPA are typically made of foam and balsa wood, covered with heat-shrinkable film, and supported by carbon fiber reinforced plastic (CFRP) pipe frameworks. In this paper, we assume this structural configuration for defining the design optimization problem.

The variables defining the main wing shape are based on dividing the right wing into $n$ segments, defined by $(n+1)$ cross-sections, as shown in Fig. \ref{fig::variables} for the case where $n=4$. The variables include $a_i$ for airfoil (cross-sectional shape) choice from DAE series\cite{drela1989xfoil}, $b_i$ for the length of each segment, $c_i$ for the chord length (wing segment length in fore-aft direction), $\alpha_i$ for the angle of attack (incidence angle of airflow), and $d_i$ for the diameter of the CFRP pipe forming the main beam. These variables are real-valued. 

\begin{figure}[t]
\centering
\begin{minipage}[b]{0.49\columnwidth}
    \centering
    \includegraphics[width=0.95\columnwidth]{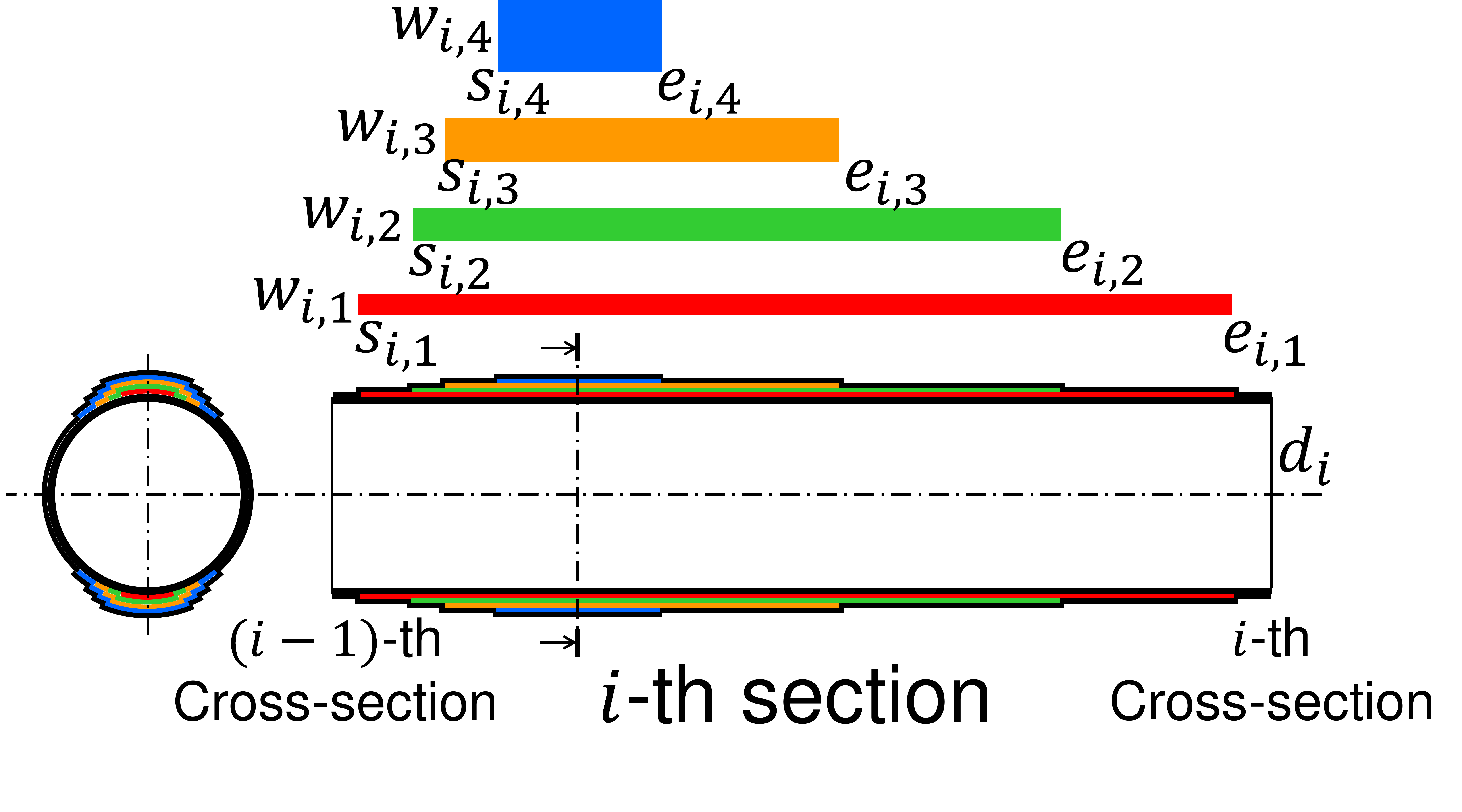}
    \caption{Variables defining the stacking sequence of CFRP pipes.}
    \label{fig::cfrp}
\end{minipage}
\begin{minipage}[b]{0.49\columnwidth}
    \centering
    \includegraphics[width=0.95\columnwidth]{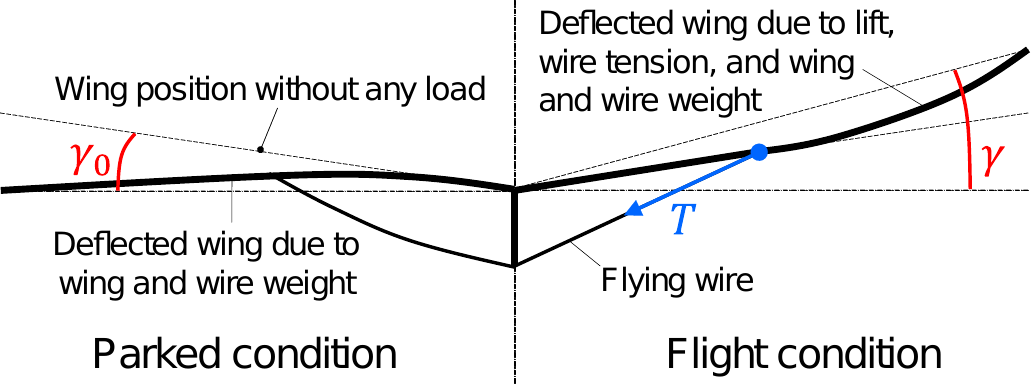}
    \caption{Relationship between wing deflection at parked and flight conditions, dihedral angle, and tension force of the flying wires.}
    \label{fig::deflection}
\end{minipage}
\end{figure}

Variables related to the stacking sequence of the CFRP pipe forming the main beam for segment $i$ are shown in Fig. \ref{fig::cfrp}. The CFRP pipe is manufactured by laminating sheets of material called prepregs, which consist of carbon fibers aligned in a specific direction, at angles (orientation angles) that match the expected loads. The CFRP pipe with a diameter $d_i$ for segment $i$ consists of a fixed set of four full-circumference laminates and $m$ variable flange laminates, which primarily support the lift force. $m$ varies depending on the optimization problem while Fig. \ref{fig::cfrp} shows an example configuration for $m=4$. In the cross-sectional view, the black area represents the full-circumference laminates, while the red, green, orange, and blue areas represent the flange laminates. 
The variables related to the stacking sequence include the starting and ending points $s_{i,j}$ and $e_{i,j}$, and the width $w_{i,j}$ for each of the $j \in {1,\cdots,m}$ flange laminates. $s_{i,j}$ and $e_{i,j}$ are non-dimensional position on the spanwise coordinate where the $(i-1)$ and $i$-th cross-sections are defined as 0 and 1, respectively. Note that for manufacturing reasons, $s_{i,j} \le s_{i,j+1}$, $e_{i,j} \ge e_{i,j+1}$, and $w_{i,j} < w_{i,j+1}$. These constraints are automatically satisfied based on the definition of the design variables.

Other variables include the dihedral angle at the wing root $\gamma_0$, the tension force $T$ of the flying wires, and the payload $W_p$. Fig. \ref{fig::deflection} illustrates the deflection curve of the main wing when viewed from the front at both parked and flight conditions. By varying $\gamma_0$, it is possible to adjust the dihedral angle at the wingtip $\gamma$ during flight, considering the need to provide clearance between the wing and the ground at the parked condition. The flying wires are introduced to relieve some of the bending moment due to lift during flight, thus reducing the weight of the inner CFRP pipes compared to when no flying wire is used. Payload $W_p$ represents the mass of cargo or passengers carried by the HPA. These three variables are included as variables in some problems.

\subsection{Design Variables}
The design variables ${\bf x}$ for the benchmark problem change with the difficulty level in the problem. Therefore, we denote the design variables at three difficulty levels as ${\bf x}_l$ for $l \in \{0, 1, 2\}$. Depending on the optimization problem and its difficulty level, a subset of variables used to define the wing is selected as design variables. However, some of the variables for defining the wing are calculated from different design variables for each difficulty level through variable transformations tailored to the difficulty.
In the definition of design variables, the higher-level design space includes the lower-level design space, allowing lower-level design variables to be transformed into higher-level design variables.
All design variables are normalized to the range [0,1] in the implementation, as designers can arbitrarily define the upper and lower bounds, which are predetermined parameters even in real-world problems.

\subsubsection{Difficulty Level $l=0$}
In this case, the design variables include segment lengths $b_i$, chord lengths $c_i$, an angle of attack at wing root $\alpha_0$, washout angles $\dot{\alpha_i}$, a dihedral angle at wing root $\gamma_0$, wire tension $T$, payload $W_p$, and the reduction coefficients for reference strains at flight and parked conditions, $r_0$ and $r_1 \in (0,1]$. Note that depending on the problem, $\gamma_0$, $T$, and $W_p$ may be treated as constants. For $l=0$, airfoil $a_i$ is considered a constant, and the diameter of the CFRP pipe $d_i$ is defined as $d_i=\min_{\tilde{y_i}} t(\tilde{y_i})c(\tilde{y_i})$, where $t(\tilde{y_i})$ and $c(\tilde{y_i})$ are functions that linearly interpolate wing thickness and chord length in the wing span direction within the $i$-th segment using $\tilde{y_i} \in [0, 1]$ as the non-dimensional spanwise coordinate.
The angle of attack $\alpha_i$ at each cross-section is transformed from $\alpha_0$ and $\dot{\alpha_i}$ as follows:

\begin{equation}
\label{alpha}
\alpha_i = \alpha_0 + \sum_{i^{\prime}=1}^{i}\dot{\alpha_{i^{\prime}}}.
\end{equation}

\noindent The variables $s_{i,j}$, $e_{i,j}$, and $w_{i,j}$ for each segment are greedily optimized to satisfy constraint $g_{1,i}^{\prime}(\tilde{y_i}, {\bf x}_l) \le 0$ with the use of $r_0$ and $r_1$:

\begin{equation}
\label{strain}
g_{1,i}^{\prime}(\tilde{y_i}, {\bf x}_l) = \frac{n_m n_s}{\epsilon_u} \max
\left[
\frac{\epsilon_0(\tilde{y_i}, {\bf x}_l)}{r_0},
\frac{\epsilon_1(\tilde{y_i}, {\bf x}_l)}{r_1}
\right] - 1,
\end{equation}

\noindent where, $n_m=1.5$, $n_s=2$, and $\epsilon_u=0.0027$ are constants representing load factor, safety factor, and reference strain. $\epsilon_0(\tilde{y_i}, {\bf x}_l)$ and $\epsilon_1(\tilde{y_i}, {\bf x}_l)$ are absolute values of compressive strains on the outer surface of CFRP pipe at $\tilde{y_i}$ at flight and parked conditions, respectively.

\subsubsection{Difficulty Level $l=1$}
In addition to the design variables from level $l=0$, airfoil $a_i$ and the diameter reduction factor $\tilde{d_i} \in (0,1]$ for the CFRP pipe are included as design variables for level $l=1$. The diameter of the CFRP pipe is defined as $d_i=\tilde{d_i} \min_{\tilde{y_i}} t(\tilde{y_i}) c(\tilde{y_i})$. The flange laminates $s_{i,j}$, $e_{i,j}$, and $w_{i,j}$ are optimized in the same way as for level $l=0$. 
However, the reduction factors $r_{0,i}$ and $r_{1,i}$ for each segment vary, and they are used as design variables instead of $r_0$ and $r_1$.

\subsubsection{Difficulty Level $l=2$}
In level $l=2$, flange laminates $s_{i,j}$, $e_{i,j}$, and $w_{i,j}$ are determined differently from level $l=1$. The design variables consist of the change in length and width between each layer of flange laminate $\dot{l}_{i,j}$ and $\dot{w}_{i,j}$, and the spanwise reference position $\xi_i \in [0,1]$. The flange laminate is defined using the following equations:

\begin{gather}
\label{length&center}
l_{i,j} = \max \left(0, \; 1 - \sum_{j^{\prime}=1}^j \dot{l}_{i,j^{\prime}} \right), \
\tilde{\xi_i} = \text{round} \left(\frac{1}{1 + \exp(-50(\xi_i - 0.5))}, \; 4 \right), \\
s_{i,j} = \tilde{\xi_i}(1 - l_{i,j}), \
e_{i,j} = \tilde{\xi_i} + l_{i,j}(1 - \tilde{\xi_i}), \
w_{i,j} = \frac{\pi d_i}{12} - \Delta w + \sum_{j^{\prime}=1}^j \dot{w}_{i,j^{\prime}},
\end{gather}

\noindent where $\Delta w = 0.002$ [m] represents the minimum increment of the laminate width. Eq. \ref{length&center} uses a sigmoid function and rounding to convert $\xi_i$ to $\tilde{\xi_i}$, where the rounding (round) operation truncates the value to four decimal places. These improve the probability that each laminate connects to one end of the CFRP pipe and leads to a reasonable design.

\subsection{Objective Functions and Constraints}
Each benchmark problem is constructed by selecting a subset of the fundamental 11 objective functions to be minimized, $f_i$, and 5 constraint conditions, $g_i\leq0$, as shown in Table \ref{tb::objcon}. The functions $P$, $D$, $V$, $\Phi$, $E$, $W_0$, and $B$ represent the required power, drag, cruise speed, twist angle at wingtip, wing efficiency, empty weight, and wing span, respectively. The variables $\delta$ and $\delta_{\text{park}}$ denote the wingtip displacement (positive upwards) at flight and parked conditions, relative to the wing root height. $\epsilon_{\text{max}}$ represents the maximum strain in CFRP pipes at both parked and flight conditions, and $\gamma$ represents the dihedral angle at wingtip at the flight condition. Additionally, $\eta=0.85\times0.95$ is the product of propeller and drivetrain efficiency, $\gamma_u=\ang{8}$ is maximum dihedral angle, $P_{\text{max}}=400$ [W] is maximum required power, and $V_{\text{min}}=7.3$ [m/s] is the minimum cruise speed. Since $V$, $E$, and $W_p$ are functions to be maximized, a negative sign is applied to $f_3$, $f_6$, and $f_{11}$ to convert them into minimization problems.

Many of the objective functions and constraints are functions of most elements in ${\bf x}_l$. However, the wing efficiency $E$ for $f_6$ depends only on the variables related to the wing shape: airfoil $a_i$, section length $b_i$, chord length $c_i$, and angle of attack $\alpha_i$. $f_8$ is the sum of $b_i$, while $f_9$ through $f_{11}$ are specific design variables themselves. Additionally, functions that have the dihedral angle at wing root $\gamma_0$ as a variable are limited to $f_4$, $g_2$, and $g_3$, and only functions $f_4$ and $g_2$ have all elements in ${\bf x}_l$ as variables.

The objective functions and constraints are evaluated by solving integral equations based on lifting line theory and differential equations for deflection based on beam-column theory. Computational grids with 50 and 301 points are used for solving these equations, respectively.

\begin{table}[t]
\caption{Fundamental functions and penalty weights for unconstrained problems.}
\begin{center}
\scalebox{0.8}{
\begin{tabular}{lcl|lcl}
\hline 
\multicolumn{1}{c}{Function} & Unit & \multicolumn{1}{c}{Weight} & \multicolumn{1}{|c}{Function} & Unit & \multicolumn{1}{c}{Weight} \\
\hline \hline
$f_1 = P({\bf x}_l) = D({\bf x}_l)V({\bf x}_l)/\eta$          & W   & $v_1=10$ & $f_9 = \alpha_0$                                              & deg & $v_9=1$ \\
$f_2 = D({\bf x}_l)$                                          & N   & $v_2=1$  & $f_{10} = T$                                                  & N   & $v_{10}=200$ \\
$f_3 = - V({\bf x}_l)$                                        & m/s & $v_3=1$  & $f_{11} = - W_p$                                              & kg  & $v_{11}=10$ \\
$f_4 = \max(|\delta({\bf x}_l)|, |\delta_{\text{park}}({\bf x}_l)|)$ & m   & $v_4=0.5$ & $g_1 = n_m n_s \epsilon_{\text{max}}({\bf x}_l)/\epsilon_u - 1$      & -   & $w_1=10$ \\
$f_5 = \Phi({\bf x}_l)$                                       & deg & $v_5=0.1$ & $g_2 = B(\sin{\gamma({\bf x}_l)} - \sin{\gamma_u})/2$         & m   & $w_2=2$ \\
$f_6 = - E({\bf x}_l)$                                        & -   & $v_6=0.1$ & $g_3 = - \delta_{\text{park}}({\bf x}_l)$                            & m   & $w_3=2$ \\
$f_7 = W_0({\bf x}_l)$                                        & kg  & $v_7=1$  & $g_4 = P({\bf x}_l) - P_{\text{max}}$                                & W   & $w_4=0.1$ \\
$f_8 = B = 2\sum_{i=1}^n b_i$                                 & m   & $v_8=1$  & $g_5 = 1 - (V({\bf x}_l)/V_{\text{min}})^3$                          & -   & $w_5=10$ \\
\hline
\end{tabular}}
\label{tb::objcon}
\end{center}
\end{table}

\subsection{Definition of Each Problem}
By selecting a subset of functions from Table \ref{tb::objcon}, we define 3 single-objective problems and 17 multi-objective problems. Each problem has three difficulty levels, resulting in a total of 60 benchmark problems. The benchmark problems proposed in this paper are denoted as HPA$MNL$-$l$, where $M$ is the number of objective functions, $N$ is the number of inequality constraints (excluding box constraints), $L$ is the problem ID, and $l$ is the difficulty level, which may be omitted if irrelevant. The wing segmentation number $n$ is a parameter that increases or decreases the problem dimension, and in this paper, we set $n=4$. $n$ should be an even number to obtain a realistic design.

In Table \ref{tb::problems}, we present the objective functions, constraints, and the design variables used for each problem, as well as the maximum number of layers in CFRP flange laminates, $m$, and the dimension of design variables $|{\bf x}_l|$ when $n=4$ for each difficulty level. The dimension of the design variables varies with $n$, $m$, and $k$ as follows: $|{\bf x}_0|=3n+4+k$, $|{\bf x}_1|=7n+3+k$, and $|{\bf x}_2|=(2m+6)n+3+k$, where $k$ is the number of variables used among $\gamma_0$, $T$, and $W_p$ listed in Table \ref{tb::problems}. Each problem is defined based on the intended use and configuration of the aircraft. For example, HPA101 and HPA102 are both designed for long-distance flights, but they differ in whether flying wires are included. In contrast, HPA103 is tailored for high-speed flight.

In unconstrained problems with $N=0$, we modify the objective functions $f_i$ as $f_i^{\prime}$ using the penalty weights $v_i$ and $w_j$ from Table \ref{tb::objcon} as:

\begin{equation}
\label{penalized}
f_i^{\prime} = f_i + v_i \sum_{j \in \mathcal{G}} w_j \max(0, g_j)
\end{equation}

\noindent
where, $\mathcal{G} \subset \{1, 2, 3, 4, 5\}$ is a set composed of constraint indices used in each problem from Table \ref{tb::problems}.

\renewcommand{\arraystretch}{0.8}
\begin{table}[t]
\caption{Benchmark problem definitions and dimensionality when $n=4$.}
\begin{center}
\scalebox{0.8}{
\begin{tabular}{c|ccccccccccc|ccccc|ccc|cccc}
\hline 
Problem & $f_1$ & $f_2$ & $f_3$ & $f_4$ & $f_5$ & $f_6$ & $f_7$ & $f_8$ & $f_9$ & $f_{10}$ & $f_{11}$ &
$g_1$ & $g_2$ & $g_3$ & $g_4$ & $g_5$ & $\gamma_0$ & $T$ & $W_p$ & $m$ & $|{\bf x}_0|$ & $|{\bf x}_1|$ & $|{\bf x}_2|$ \\
\hline \hline
HPA131, HPA101 &  & \checkmark & &  &  &  &  &  &  &  &  & 
\checkmark & \checkmark & \checkmark &  &  &  & \checkmark &  & 10 & 17 & 32 & 108 \\
HPA142, HPA102 & \checkmark &  &  &  &  &  &  &  &  &  &  & 
\checkmark & \checkmark & \checkmark &  & \checkmark & \checkmark &  &  & 20 & 17 & 32 & 188 \\
HPA143, HPA103 &  &  & \checkmark &  &  &  &  &  &  &  &  & 
\checkmark & \checkmark & \checkmark & \checkmark &  & \checkmark &  &  & 20 & 17 & 32 & 188 \\
\hline
HPA241, HPA201 & \checkmark &  & \checkmark &  &  &  &  &  &  &  &  & 
\checkmark & \checkmark & \checkmark & \checkmark &  & \checkmark & \checkmark &  & 10 & 18 & 33 & 109 \\
HPA222, HPA202 & \checkmark &  &  & \checkmark &  &  &  &  &  &  &  & 
\checkmark &  &  &  & \checkmark &  &  &  & 20 & 16 & 31 & 187 \\
HPA233, HPA203 &  & \checkmark &  &  &  &  &  &  &  &  & \checkmark & 
\checkmark & \checkmark & \checkmark &  &  & \checkmark & \checkmark & \checkmark & 20 & 19 & 34 & 190 \\
HPA244, HPA204 &  &  & \checkmark &  & \checkmark &  &  &  &  &  &  & 
\checkmark & \checkmark & \checkmark & \checkmark &  & \checkmark & \checkmark &  & 20 & 18 & 33 & 189 \\
HPA245, HPA205 &  &  & \checkmark &  &  &  &  &  &  & \checkmark &  & 
\checkmark & \checkmark & \checkmark & \checkmark &  & \checkmark & \checkmark &  & 20 & 18 & 33 & 189 \\
\hline
HPA341, HPA301 & \checkmark &  & \checkmark &  &  & \checkmark &  &  &  &  &  & 
\checkmark & \checkmark & \checkmark & \checkmark &  & \checkmark & \checkmark &  & 10 & 18 & 33 & 109 \\
HPA322, HPA302 & \checkmark &  &  & \checkmark & \checkmark &  &  &  &  &  &  & 
\checkmark &  &  &  & \checkmark &  &  &  & 20 & 16 & 31 & 187 \\
HPA333, HPA303 &  & \checkmark &  &  &  &  &  &  & \checkmark &  & \checkmark & 
\checkmark & \checkmark & \checkmark &  &  & \checkmark & \checkmark & \checkmark & 20 & 19 & 34 & 190 \\
HPA344, HPA304 &  & \checkmark & \checkmark &  & \checkmark &  &  &  &  &  &  & 
\checkmark & \checkmark & \checkmark & \checkmark &  & \checkmark & \checkmark &  & 20 & 18 & 33 & 189 \\
HPA345, HPA305 & \checkmark &  & \checkmark &  &  &  &  &  &  & \checkmark &  & 
\checkmark & \checkmark & \checkmark & \checkmark &  & \checkmark & \checkmark &  & 20 & 18 & 33 & 189 \\
\hline
HPA441, HPA401 &  & \checkmark & \checkmark &  & \checkmark &  & \checkmark &  &  &  &  & 
\checkmark & \checkmark & \checkmark & \checkmark &  & \checkmark & \checkmark &  & 20 & 18 & 33 & 189 \\
HPA422, HPA402 & \checkmark &  &  & \checkmark & \checkmark &  &  & \checkmark &  &  &  & 
\checkmark &  &  &  & \checkmark &  &  &  & 20 & 16 & 31 & 187 \\
HPA443, HPA403 &  &  & \checkmark &  &  & \checkmark & \checkmark &  &  & \checkmark &  & 
\checkmark & \checkmark & \checkmark & \checkmark &  & \checkmark & \checkmark &  & 20 & 18 & 33 & 189 \\
\hline
HPA541, HPA501 &  & \checkmark & \checkmark &  & \checkmark & \checkmark & \checkmark &  &  &  &  & 
\checkmark & \checkmark & \checkmark & \checkmark &  & \checkmark & \checkmark &  & 20 & 18 & 33 & 189 \\
HPA542, HPA502 & \checkmark &  & \checkmark &  &  &  &  &  & \checkmark & \checkmark & \checkmark & 
\checkmark & \checkmark & \checkmark & \checkmark &  & \checkmark & \checkmark & \checkmark & 20 & 19 & 34 & 190 \\
\hline
HPA641, HPA601 & \checkmark &  & \checkmark &  & \checkmark & \checkmark & \checkmark & \checkmark &  &  &  & 
\checkmark & \checkmark & \checkmark & \checkmark &  & \checkmark & \checkmark &  & 20 & 18 & 33 & 189 \\
\hline
HPA941, HPA901 & \checkmark &  & \checkmark &  & \checkmark & \checkmark & \checkmark & \checkmark & \checkmark & \checkmark & \checkmark & 
\checkmark & \checkmark & \checkmark & \checkmark &  & \checkmark & \checkmark & \checkmark & 20 & 19 & 34 & 190 \\
\hline
\end{tabular}}
\label{tb::problems}
\end{center}
\end{table}
\renewcommand{\arraystretch}{1}

\section{Numerical Experiment}
\subsection{Setups}
We conducted experiments on 60 unconstained HPA problems with Eq. \ref{penalized} as objective functions because most widely available EAs and BO methods lack sufficient support for handling constraints. 

\subsubsection{Single-Objective Problems}
In the single-objective problem, we verify whether the primary objective functions exhibit smooth landscapes and moderate multimodality. 
We employed algorithms from the black-box optimization libraries pymoo\cite{pymoo}, BoTorch\cite{botorch}, and Optuna\texttrademark \cite{optuna}. From pymoo, we utilized four EAs, namely covariance matrix adaptation evolution strategy (CMA-ES)\cite{cmaes}, particle swarm optimization (PSO), differential evolution (DE), and genetic algorithm (GA), with default settings for parameters except for the population size. From BoTorch, we utilized BO methods, namely GP model with a Mat\'ern-5/2 kernel and expected improvement (GP-EI), and trust region BO (TuRBO)\cite{turbo}. Optuna\texttrademark \ provided the tree-structured Parzen estimator (TPE)\cite{tpe}.

In all algorithms, the population size and the number of initial sample points were set to 20, the function evaluation budget was set to 1,000, the number of independent runs was set to 11. The batch size (number of additional sample points at each iteration) of GP-EI, TuRBO, and TPE was set to 1. Other parameters were set to the default values in each library.

\subsubsection{Multi-Objective Problems}
In the multi-objective HPA problem, we reveal the PF shapes, the characteristics of non-dominated solutions (NDSs) achievable by existing algorithms, and the differences from conventional benchmark problems.
We employed six algorithms from pymoo: non-dominated sorting genetic algorithm II (NSGA-II) \cite{nsga2} and III (NSGA-III) \cite{nsga3}, multi-objective EA based on decomposition (MOEA/D) \cite{moead}, S-metric selection evolutionary multi-objective optimisation algorithm (SMS-EMOA) \cite{smsemoa}, reference vector guided EA (RVEA) \cite{rvea}, and adaptive geometry estimation based many-objective EA II (AGE-MOEA-II) \cite{agemoea2}. In all algorithms, the population size was set to $40M$, where $M$ is the number of objective functions, and the function evaluation budget was set to 72,000 for problems with $l=0,1$, and 216,000 for problems with $l=2$. Reference vectors for NSGA-III, MOEA/D, and RVEA are generated by the Riesz s-Energy method \cite{s-energy}, and the number of vectors is the same as the population size. As pointed out by He et al. \cite{he2023}, objective function normalization was employed for MOEA/D, SMS-EMOA, and RVEA to improve their performance. 
Other parameters were set to the default values in pymoo.

We conducted 11 independent runs and calculated the inverted generational distance plus ($\text{IGD}^+$) \cite{ishibuchi2015modified} for the normalized objective functions. The reference points for $\text{IGD}^+$ were selected from NDSs obtained by hot-started runs of the six algorithms under the same settings as the cold-started runs, except for the initialization, where the initial population consisted of NDSs extracted from the external archive of solutions obtained across all cold-started runs, encompassing all problems, difficulty levels, and algorithms. The normalization also used the utopia and nadir points derived from the reference points.

As conventional benchmark problems, we use the RE \cite{tanabe2020easy} problems, and 30-dimensional DTLZ1-7\cite{dtlz} and Minus-DTLZ1-7\cite{ishibuchi2016performance} problems ($M=2, 3, 6$). The experimental settings are identical to those for the HPA problems. For the RE problems, DTLZ5, 6 ($M=6$), and Minus-DTLZ5–7, we used a subset of NDSs from all cold-started runs of six algorithms as reference points for $\text{IGD}^+$.

\subsection{Pareto Front Shapes}
\begin{figure}[t]
\centerline{\includegraphics[width=0.95\columnwidth]{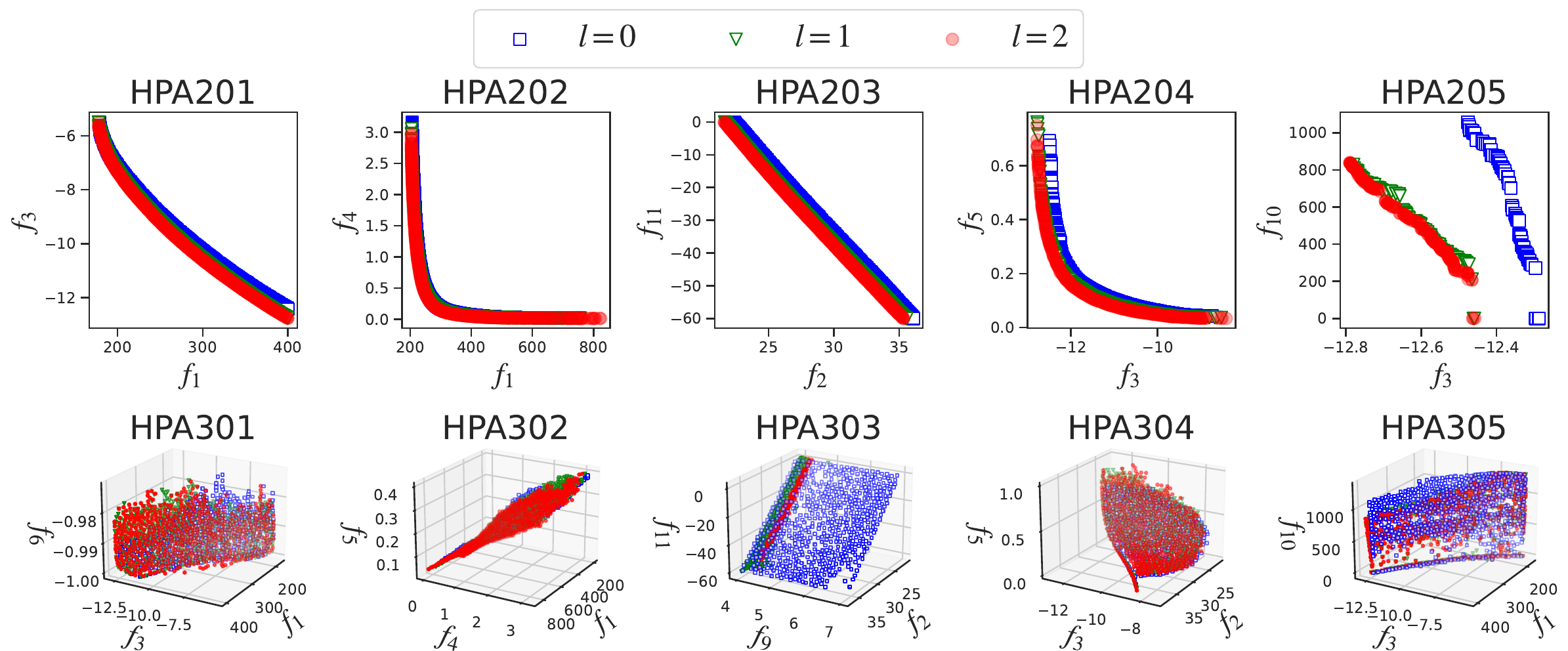}}
\vspace{-4mm}
\caption{Pareto front shapes with reference points for $\text{IGD}^+$ in each difficulty level.}
\label{fig::pf}
\end{figure}

The $\text{IGD}^+$ reference points for two-objective and three-objective problems are shown in Fig. \ref{fig::pf}. While many problems have convex PF, HPA203 and 303 have linear PF and HPA205 and 305 have concave PF in particular cross-sections. When HPA305 is sliced parallel to the $f_{10}$ axis, a PF shape similar to HPA205 appears. HPA304 forms an inverted triangular PF. In HPA302, there is a strong positive correlation between $f_4$ and $f_5$, causing a partially degenerated PF near the minimum values of these objectives. In HPA problems, the PFs of higher-difficulty level dominate those of lower-difficulty level since the design space of higher-difficulty level includes that of lower-difficulty levels. 

\subsection{Results}
\subsubsection{Single-Objective Problems}
\begin{figure}[t]
\centerline{\includegraphics[width=0.95\columnwidth]{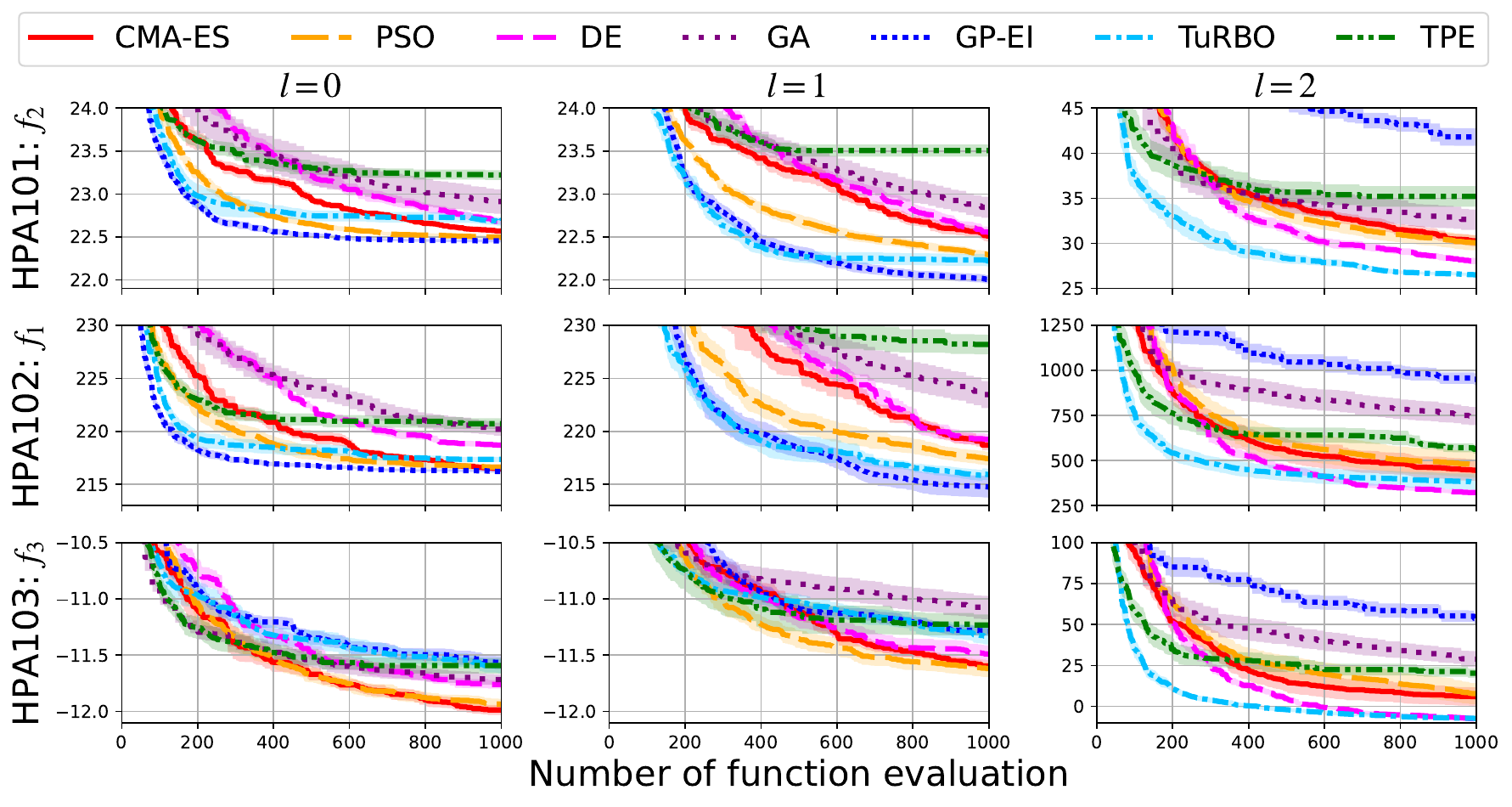}}
\vspace{-4mm}
\caption{Histories of mean objective function values and standard errors in the single-objective HPA problems.}
\label{fig::history}

\vspace{4mm}
\centerline{\includegraphics[width=0.95\columnwidth]{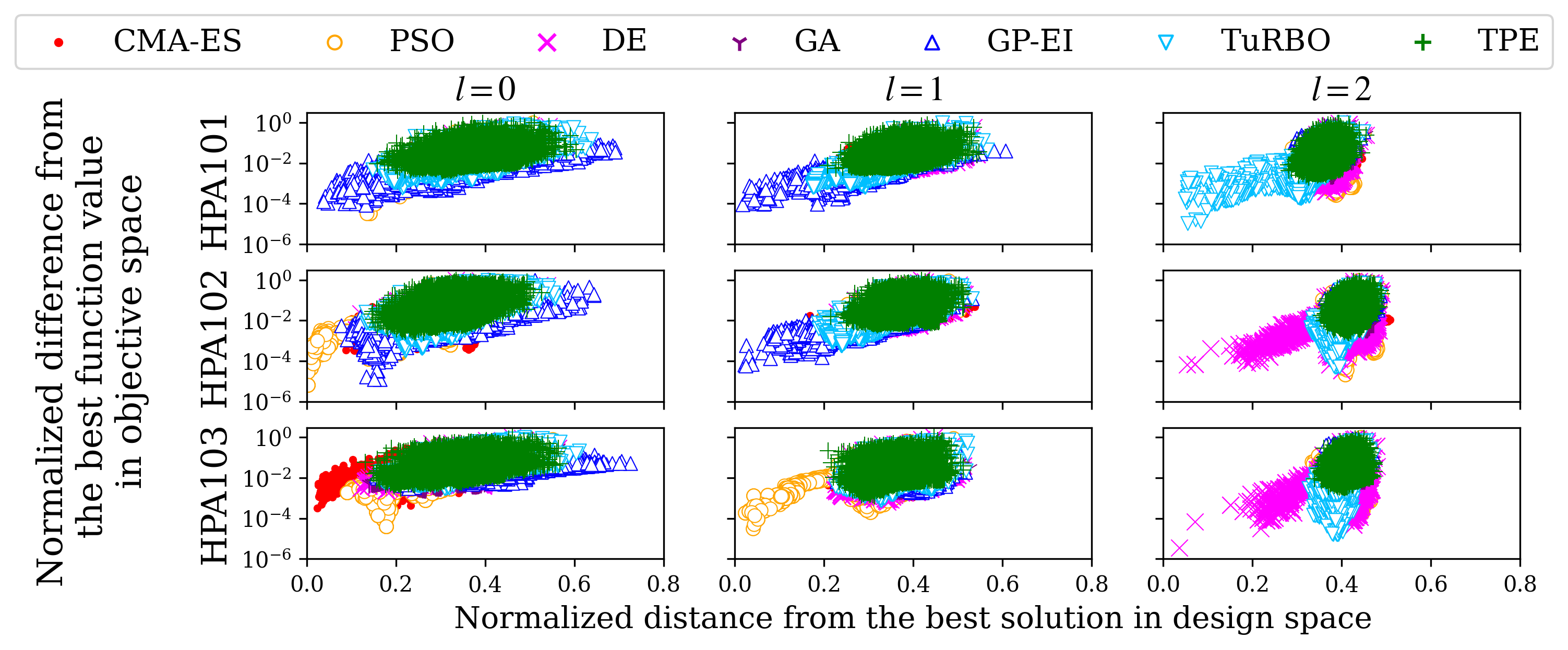}}
\vspace{-4mm}
\caption{Fitness-distance scatter plots for the single-objective HPA problems.}
\label{fig::fdc}
\end{figure}

Fig. \ref{fig::history} presents the averaged results over 11 runs of the best objective function values and their standard errors at each point up to 1,000 function evaluations. 
In HPA101-0,1 and HPA102-0,1, algorithms such as TuRBO, GP-EI, and PSO demonstrated consistently high performance across a broad range. These algorithms are widely utilized in computationally expensive problems \cite{namura2019multipoint, sun2017surrogate}. The effective application of algorithms incorporating GP models indicates that HPA problems likely exhibit smooth
landscapes, as intended. Conversely, PSO, CMA-ES, and DE achieved favorable outcomes in HPA103-0,1, while GP-EI and TuRBO showed only moderate success. This suggests that these problems might pose challenges for surrogate model approximation. For problems at difficulty level $l=2$, TuRBO, specifically designed for high-dimensional problems, generally outperformed others, with DE following. 

To validate the moderate multimodality of the HPA problems, we generated fitness-distance scatter plots\cite{jones1995fitness} for each problem, as shown in Fig. \ref{fig::fdc}, using all solutions. The vertical axis represents the difference between the normalized objective function value of each solution and that of the best solution, while the horizontal axis shows the normalized distance between each solution and the best solution. The results reveal that each HPA problem exhibits a large global valley but also contains multiple local optima. Specifically, HPA101-0,1, HPA102-0,1, and HPA103-0 present relatively simple structures, whereas the problems with $l=2$ and HPA103-1 have many local optima or deceptive structures.

\subsubsection{Multi-Objective Problems}

\renewcommand{\arraystretch}{0.8}
\begin{table}[t]
\caption{Comparison of mean $\text{IGD}^+$ across 11 runs. Values separated by `/' indicate: the number of problems with the best mean $\text{IGD}^+$, those with no significant difference (Wilcoxon rank-sum test, $p=0.05$) from the best, and the total number of problems. The best algorithm (sum of $a$ and $b$ in `$a/b/c$') for each problem set is in bold.}
\begin{center}
\scalebox{0.8}{
\begin{tabular}{cc|cccccc}
\hline 
Problem & $M$ & NSGA-II & NSGA-III & MOEA/D & SMS-EMOA & RVEA & AGE-MOEA-II \\
\hline \hline 
HPA & 2 & \textbf{8/6/15} & 1/2/15 & 0/1/15 & 2/6/15 & 3/2/15 & 1/12/15 \\
HPA & 3 & 3/5/15 & 1/3/15 & 0/0/15 & \textbf{9/4/15} & 0/1/15 & 2/3/15 \\
HPA & 4-9 & 2/3/21 & 3/6/21 & 0/0/21 & 0/1/21 & 2/0/21 & \textbf{14/3/21} \\
HPA & Total & 13/14/51 & 5/11/51 & 0/1/51 & 11/11/51 & 5/3/51 & \textbf{17/18/51} \\
\hline
RE & 2 & 2/0/5 & 0/0/5 & 0/0/5 & \textbf{3/0/5} & 0/0/5 & 0/0/5 \\
RE & 3 & 2/0/7 & \textbf{2/2/7} & 0/0/7 & 2/0/7 & 0/0/7 & 1/0/7 \\
RE & 4-9 & 0/0/4 & 0/0/4 & 0/0/4 & \textbf{2/0/4} & 0/0/4 & \textbf{2/0/4} \\
RE & Total & 4/0/16 & 2/2/16 & 0/0/16 & \textbf{7/0/16} & 0/0/16 & 3/0/16 \\
\hline 
DTLZ & 2 & \textbf{3/1/7} & 0/3/7 & 2/0/7 & 1/1/7 & 1/0/7 & 0/3/7 \\
DTLZ & 3 & 2/0/7 & \textbf{3/0/7} & 0/1/7 & 1/1/7 & 0/2/7 & 1/1/7 \\
DTLZ & 6 & 0/0/7 & 1/0/7 & 2/1/7 & 0/0/7 & \textbf{3/1/7} & 1/0/7 \\
DTLZ & Total & 5/1/21 & \textbf{4/3/21} & 4/2/21 & 2/2/21 & \textbf{4/3/21} & 2/4/21 \\
\hline
Minus-DTLZ & 2 & \textbf{4/0/7} & 0/0/7 & 0/0/7 & 2/0/7 & 0/0/7 & 1/1/7 \\
Minus-DTLZ & 3 & \textbf{4/0/7} & 1/1/7 & 0/0/7 & 1/0/7 & 0/0/7 & 1/1/7 \\
Minus-DTLZ & 6 & \textbf{4/1/7} & 0/0/7 & 1/0/7 & 0/0/7 & 1/1/7 & 1/1/7 \\
Minus-DTLZ & Total & \textbf{12/1/21} & 1/1/21 & 1/0/21 & 3/0/21 & 1/1/21 & 3/3/21 \\
\hline
\end{tabular}}
\label{tb::igd_best}
\end{center}
\end{table}

\begin{table}[t]
\caption{Geometric mean of mean $\text{IGD}^+$ and mean evaluation time per solution in HPA problems. The smallest $\text{IGD}^+$ in each problem set is highlighted in bold.}
\begin{center}
\scalebox{0.8}{
\begin{tabular}{cc|cccccc|c}
\hline 
$M$ & $l$ & NSGA-II & NSGA-III & MOEA/D & SMS-EMOA & RVEA & AGE-MOEA-II & Time [s] \\
\hline \hline
2 & 0 & \textbf{0.01863}  & 0.02785  & 0.05085  & 0.02201  & 0.03101  & 0.02108  & 0.12108  \\
2 & 1 & \textbf{0.02957}  & 0.04429  & 0.11759  & 0.03294  & 0.04911  & 0.03351  & 0.12228  \\
2 & 2 & \textbf{0.21735}  & 0.25177  & 0.31389  & 0.23509  & 0.27890  & 0.22171  & 0.03230  \\
\hline
3 & 0 & 0.01400  & 0.01471  & 0.03950  & \textbf{0.01019}  & 0.02526  & 0.01464  & 0.12073  \\
3 & 1 & 0.02031  & 0.01949  & 0.08164  & \textbf{0.01234}  & 0.03362  & 0.01888  & 0.12191  \\
3 & 2 & \textbf{0.09508}  & 0.10224  & 0.21946  & 0.09585  & 0.17314  & 0.10319  & 0.03200  \\
\hline
4-9 & 0 & 0.03078  & 0.02168  & 0.05425  & 0.02649  & 0.03630  & \textbf{0.01947}  & 0.12860  \\
4-9 & 1 & 0.04746  & 0.03094  & 0.10488  & 0.03458  & 0.04864  & \textbf{0.02765}  & 0.13146  \\
4-9 & 2 & 0.11569  & 0.12582  & 0.32590  & 0.16379  & 0.20113  & \textbf{0.10993}  & 0.03356  \\

\hline
\end{tabular}}
\label{tb::igd}
\end{center}
\end{table}
\renewcommand{\arraystretch}{1}

\begin{figure}[t]
\centerline{
\includegraphics[width=0.95\columnwidth]{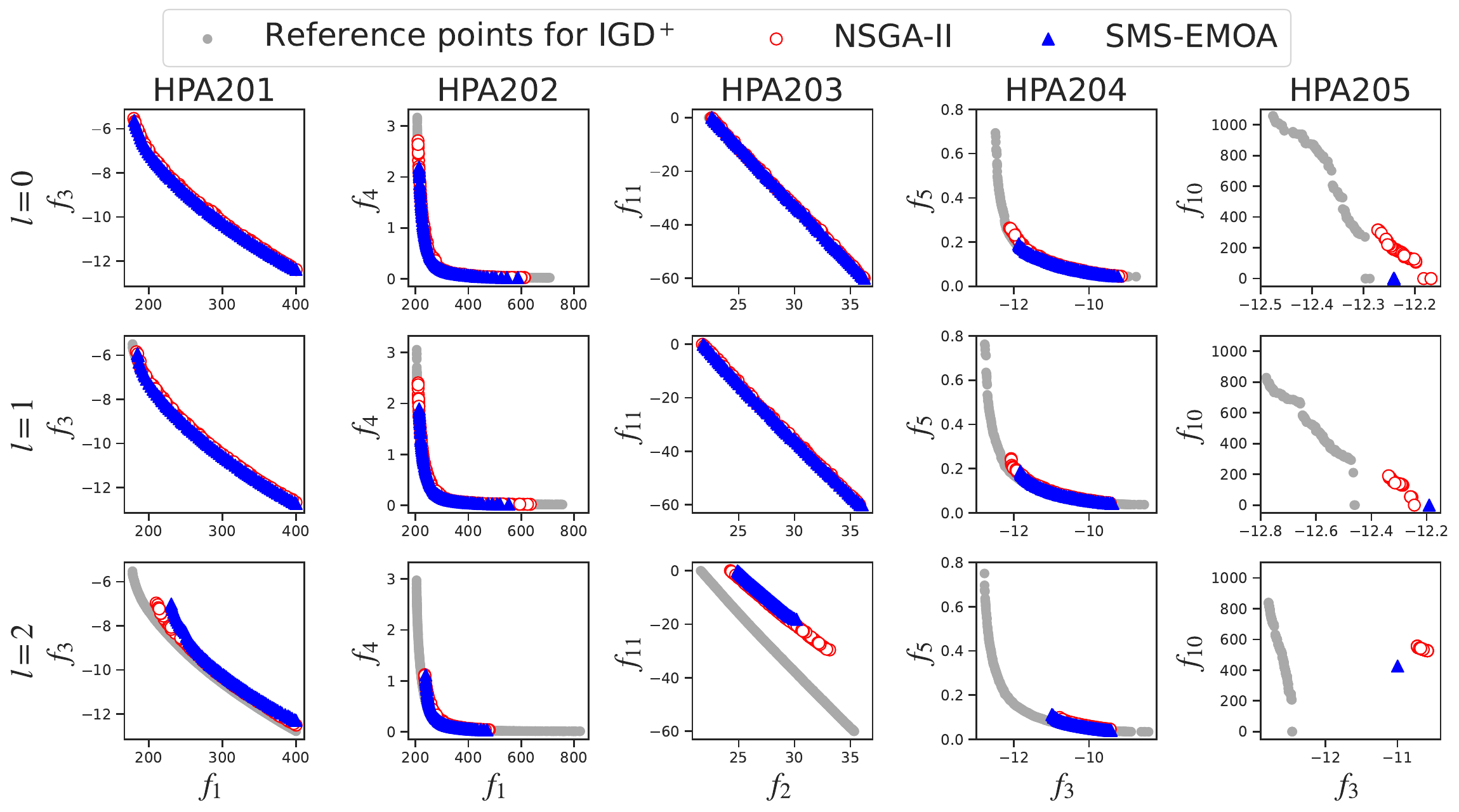}
}
\vspace{-4mm}
\caption{NDSs at the median $\text{IGD}^+$ runs in the two-objective HPA problems.}
\label{fig::nds2d}
\end{figure}

\begin{figure}[t]
\centerline{
\includegraphics[width=0.95\columnwidth]{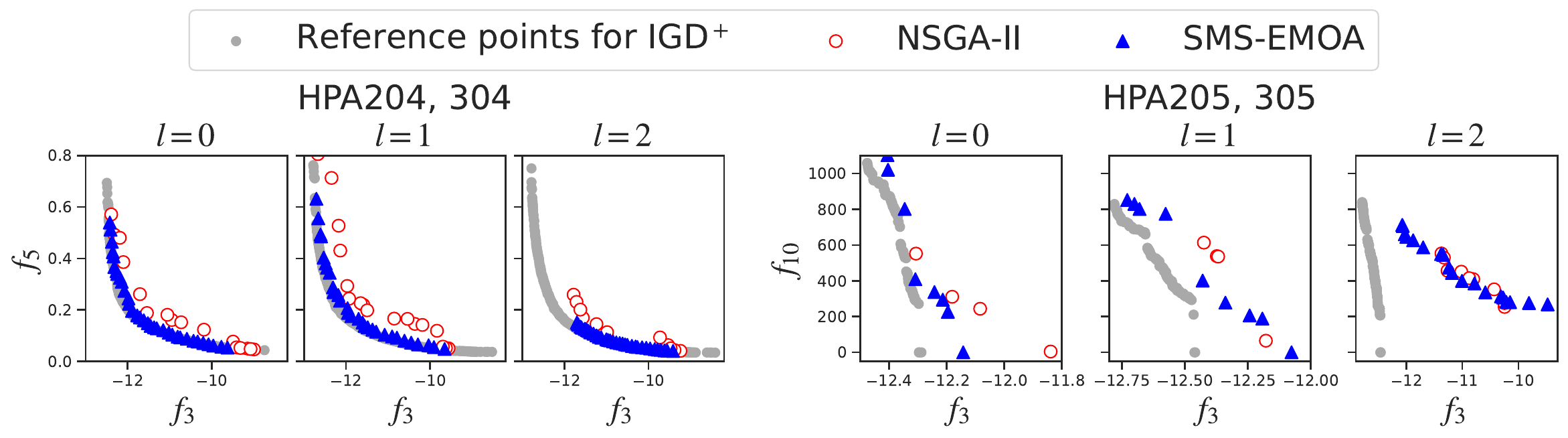}
}
\vspace{-4mm}
\caption{NDSs at the median $\text{IGD}^+$ runs in HPA304 and 305, projected into corresponding two-objective problems.}
\label{fig::nds3d}
\end{figure}

Table \ref{tb::igd_best} presents the comparison results of the mean $\text{IGD}^+$ for each benchmark problem. For HPA problems, Table \ref{tb::igd} also shows the geometric mean of mean $\text{IGD}^+$ across problems categorized by their $M$ and $l$. 
AGE-MOEA-II achieves the smallest $\text{IGD}^+$ across 17 HPA problems, including 14 many-objective problems. It also shows comparable performance with no significant difference from the best across 18 problems. NSGA-II (smallest $\text{IGD}^+$ across 13 problems) and SMS-EMOA (smallest $\text{IGD}^+$ across 11 problems) exhibit superior performance in two and three-objective problems, respectively. 
On the other hand, MOEA/D exhibits the largest $\text{IGD}^+$ in most HPA problems although objective function normalization was employed. RVEA also shows modest performance while it achieves the smallest $\text{IGD}^+$ across 5 HPA problems. 

In real-world-inspired RE problems, SMS-EMOA, NSGA-II, and AGE-MOEA-II achieve favorable results as observed in the HPA problems, suggesting certain similarities between the two test suites. However, noticeable differences exist in the algorithm rankings within each problem set. Since RE problems tend to converge more easily to the PF, algorithms emphasizing diversity preservation over convergence demonstrate better performance.
These algorithms also exhibit superior performance in Minus-DTLZ problems. However, unlike the HPA and RE problems, NSGA-II consistently outperforms the other algorithms regardless of the number of objectives.
DTLZ problems, on the other hand, show trends distinct from those observed in the HPA, RE, and Minus-DTLZ problems. In the DTLZ problems, NSGA-III, RVEA, and MOEA/D, which perform moderately in the other test suites, achieve strong performance. Conversely, AGE-MOEA-II and SMS-EMOA, which excel in the HPA problems, do not demonstrate superior performance in the DTLZ problems. This highlights the unique characteristics of the HPA problems compared to conventional benchmark problems.

In HPA problems, the $\text{IGD}^+$ values for $l=2$ are significantly higher than for $l=1$. Despite both $l=1$ and $l=2$ problems having nearly identical PFs, as shown in Fig. \ref{fig::pf}, it suggests that the six algorithms did not obtain well-converged and diverse NDSs for the $l=2$ problems due to their high dimensionality. Fig. \ref{fig::nds2d} shows NDSs obtained by NSGA-II and SMS-EMOA at the median $\text{IGD}^+$ runs in the two-objective problems for each difficulty level. Both convergence and diversity are not sufficient for the $l=2$ problems.

Additionally, HPA204 and 205 are hard to optimize even for $l=0,1$ especially near the extreme solution for $f_3$ because solutions around the $f_3$ minimum have different features in design variables from those mapping to other regions of the PF. Such solutions were easily obtained by extending the number of objective functions from HPA204 to HPA304 and from HPA205 to HPA305. Fig. \ref{fig::nds3d} shows NDSs for HPA204 and 205 obtained by NSGA-II and SMS-EMOA at the median $\text{IGD}^+$ runs in HPA304 and 305, respectively. Solving two-objective problems as if they were three-objective problems, similar to the approach of transforming a single-objective problem into a multi-objective one as discussed in \cite{knowles2001}, improves the diversity of NDSs as compared to the case shown in Fig. \ref{fig::nds2d}.

Table \ref{tb::igd} also includes the mean evaluation time of objective functions when generating 1,000 random solutions for each problem. The CPU used for calculations was an AMD Ryzen 5 3400G 3.70GHz. In all problems, the evaluation time per solution is within 0.15 seconds, and 10,000 solutions can be evaluated within 25 minutes. For problems with $l=0, 1$, where multiple strength evaluations are performed during the flange laminate optimization, the evaluation time is approximately four times longer than that of $l=2$ problems.

\section{Conclusion}
We proposed a set of single and multi-objective optimization benchmark problems focusing on the design of human-powered aircraft. Each problem can be adjusted across three difficulty levels by modifying the definition of the design variables, and the problem dimensionality can be further increased or decreased through a wing segmentation parameter. While the original benchmark problems include constraints, we also defined unconstrained versions using penalty methods. In the multi-objective problems, the Pareto fronts display diverse shapes, including the inverted triangle commonly observed in real-world problems. By applying representative evolutionary algorithms and Bayesian optimization methods to these benchmark problems, we confirmed that they likely exhibit moderate multimodality, a typical feature of engineering design problems. Moreover, existing algorithms rarely produce well-converged and diverse non-dominated solutions at the highest difficulty level. Regarding evaluation time, each solution can be evaluated in under 0.15 seconds, making the suite suitable for benchmarking multiple algorithms over tens of thousands of function evaluations.

\begin{credits}
\subsubsection{\discintname}
The authors have no competing interests to declare that are
relevant to the content of this article.
\end{credits}

%
%
\bibliographystyle{splncs04}
\bibliography{reference}

\clearpage
\appendix
\section*{Appendices}
\section{Requirements for Creating Benchmark Problems}
\label{apx:requirement}
When creating benchmark problems, it is important to consider not only their similarity to real-world problems but also their convenience for benchmarking purposes. Below, we outline the requirements for creating benchmark problems in this paper:

\begin{enumerate}
\item Design problems to be scalable, allowing for variations in difficulty level and the dimensionality of design variables.
\item Ensure that the design space with low-difficulty level is a subset of that with high-difficulty level, enabling the comparison of optimal solutions across different difficulty levels.
\item Provide multi-objective problems with diverse Pareto front (PF) shapes, including inverted triangle PF.
\item Make use of penalty methods to transform constrained real-world problems into unconstrained ones, enabling their use as unconstrained benchmark problems.
\item Ensure similarity to real-world engineering design problems by calculating objective functions and constraints based on governing equations.
\item Ensure moderate multimodality where surrogate models are applicable
\item Ensure that each solution evaluation takes less than 0.2 [s], allowing sufficient number of function evaluations for benchmark.
\end{enumerate}

\section{Flange Laminate Optimization}
\label{apx:flange}
For the HPA problems with difficulty level $l \in \{0, 1\}$, the flange laminates of the CFRP pipes are optimized using Algorithm \ref{alg::layer}. $\Delta w$ represents the minimum increment of the laminate width, set to 2 [mm] in this paper for manufacturing convenience. By applying Algorithm \ref{alg::layer} in order from the main spar near the wingtip ($i=n$), $s_{i,j}$, $e_{i,j}$, and $w_{i,j}$ are determined to satisfy the strength constraint. To ensure that the design space for $l=0,1$ is included in that for $l=2$, lines 17-26 in Algorithm \ref{alg::layer} is applied to align one end of each laminate with the section boundary (pipe end). Smaller values of $r_0$ and $r_1$ increase the stiffness of the main spar and make it easier to find solutions that satisfy deflection constraints.

\begin{algorithm}[t]
    \caption{Flange laminate optimization}
    \label{alg::layer}
    \begin{algorithmic}[1]
    \REQUIRE $i, \ m, \ r_0, \ r_1, \ \Delta w, \ {\bf x}_l$
    \ENSURE $s_{i,j}, \ e_{i,j}, \ w_{i,j} \ (j=1,2,\cdots,m)$
    \STATE Initialize $i$-th beam: $s_{i,j}=0, \ e_{i,j}=1, \ w_{i,1}=\pi d_i/12, \ w_{i,j+1}=w_{i,j} + \Delta w, \ \Delta \phi=0$
    \STATE Evaluate $g_{1,i}^{\prime}(\tilde{y_i}, {\bf x}_l)$
    \WHILE{$\exists \tilde{y_i} \in [0, 1], \ g_{1,i}^{\prime}(\tilde{y_i}, {\bf x}_l) > 0$}
        \STATE $\Delta \phi \leftarrow \Delta \phi + \pi/180$
        \STATE $w_{i,1}=\left(\pi/6 + \Delta \phi \right)d_i/2, \ w_{i,j+1}=w_{i,j} + \Delta w$
        \STATE Evaluate $g_{1,i}^{\prime}(\tilde{y_i}, {\bf x}_l)$
    \ENDWHILE
    \FOR{$j=m:1$}
        \STATE Evaluate $g_{1,i}^{\prime}(\tilde{y_i}, {\bf x}_l)$ with $e_{i,j}=0$
        \STATE $s_{i,j} = \min(\tilde{y_i} \in [0,1] \mid g_{1,i}^{\prime}(\tilde{y_i}, {\bf x}_l) > 0)$ \STATE $e_{i,j} = \max(\tilde{y_i} \in [0,1] \mid g_{1,i}^{\prime}(\tilde{y_i}, {\bf x}_l) > 0)$
        \IF{$j<m$}
            \STATE $s_{i,j} \leftarrow \min(s_{i,j}, s_{i,j+1})$
            \STATE $e_{i,j} \leftarrow \max(e_{i,j}, e_{i,j+1})$
        \ENDIF
    \ENDFOR
    \STATE $S_i = \sum_{j=1}^m s_{i,j}$, $E_i = \sum_{j=1}^m (1-e_{i,j})$
    \IF{$S_i E_i>0$}
        \FOR{$j=m:1$}
            \IF{$S_i > E_i$}
                \STATE $e_{i,j} = 1$
            \ELSE
                \STATE $s_{i,j} = 0$
            \ENDIF
        \ENDFOR
    \ENDIF
    \end{algorithmic}
\end{algorithm}

\section{Design Space}
\label{apx:domain}
The lower and upper bounds of the design variables are shown in Table \ref{tb::bound} where $g$ [m/s] is the gravitational acceleration, while all design variables have been normalized to the range [0,1] in the implementation.

\section{Comparison between HPA problems and conventional benchmark problems}
\label{apx:comparison}
In addition to the results shown in Table \ref{tb::igd_best}, Table \ref{tb::igd_best_all} presents a comparison of the performance of the six algorithms on both WFG and Minus-WFG problems. For these problem sets, SMS-EMOA achieves the best results, and AGE-MOEA-II performs well in the six-objective problems. On the three-objective Minus-WFG problems, NSGA-III exhibits the best performance, which is consistent with the results observed in RE and DTLZ problems. Although AGE-MOEA-II, NSGA-II, and SMS-EMOA show comparable performance on the HPA problems, the performance of NSGA-II is more modest on the WFG and Minus-WFG problems, indicating a difference in behavior across these test suites.

\section{$\text{IGD}^+$ for Each HPA Problem}
\label{apx:igd}
Table \ref{tb::igd_detail} presents the mean $\text{IGD}^+$ in 11 runs for each algorithm and the mean evaluation time per solution when generating 1,000 random solutions for each problem. To evaluate the overall performance of the algorithms, we computed a geometric mean of mean $\text{IGD}^+$. AGE-MOEA-II and SMS-EMOA show the best two geometric mean of $\text{IGD}^+$. 
On the other hand, MOEA/D exhibits the largest $\text{IGD}^+$ in most problems, and RVEA shows modest performance.

\renewcommand{\arraystretch}{0.8}
\begin{table}[h]
\caption{Upper and Lower Bounds for Design Variables.}
\begin{center}
\scalebox{0.8}{
\begin{tabular}{cc|cccc||cc|cccc}
\hline 
$l$   & ${\bf x}_l$ & Dimension & Min. & Max. & Unit & $l$   & ${\bf x}_l$ & Dimension & Min. & Max. & Unit\\
\hline \hline
0,1,2 & $b_i$                    & $n$   & $8/n$ & $20/n$ & m   & 0     & $r_0$                    & 1     & 0.01  & 1      & - \\
      & $c_i$                    & $n+1$ & 0.4   & 1.5    & m   &       & $r_1$                    & 1     & 0.01  & 1      & - \\
      & $\alpha_0$               & 1     & 4     & 7      & deg & 1     & $r_{0,i}$                & $n$   & 0.01  & 1      & - \\
      & $\dot{\alpha_i}$         & $n$   & $-8/n$& 0      & deg &       & $r_{1,i}$                & $n$   & 0.01  & 1      & - \\
      & $\gamma_0$               & 1     & 0     & 4      & deg & 2     & $\dot{l}_{i,j}$          & $mn$  & 0     & 1      & - \\
      & $T$                      & 1     & 0     & $135g$ & N   &       & $\dot{w}_{i,j}$          & $mn$  & 0.002 & 0.26   & m \\
      & $W_p$                    & 1     & 0     & 60     & kg  &       & $\xi_i$                  & $n$   & 0     & 1      & - \\
\hline
1,2   & $a_i$                    & $n+1$ & 0     & 3      & - \\
      & $\tilde{d_i}$            & $n$   & 0.8   & 1      & - \\
\hline
\end{tabular}}
\label{tb::bound}
\end{center}
\vspace{3mm}

\renewcommand{\arraystretch}{1}
\caption{Comparison of mean $\text{IGD}^+$ for HPA problems and conventional benchmark problems including WFG and Minus-WFG problems. Values separated by `/' indicate: the number of problems with the best mean $\text{IGD}^+$, those with no significant difference (Wilcoxon rank-sum test, $p=0.05$) from the best, and the total number of problems. The best algorithm (sum of $a$ and $b$ in `$a/b/c$') for each problem set is in bold.}
\begin{center}
\scalebox{0.8}{
\begin{tabular}{cc|cccccc}
\hline 
Problem & $M$ & NSGA-II & NSGA-III & MOEA/D & SMS-EMOA & RVEA & AGE-MOEA-II \\
\hline \hline 
HPA & 2 & \textbf{8/6/15} & 1/2/15 & 0/1/15 & 2/6/15 & 3/2/15 & 1/12/15 \\
HPA & 3 & 3/5/15 & 1/3/15 & 0/0/15 & \textbf{9/4/15} & 0/1/15 & 2/3/15 \\
HPA & 4-9 & 2/3/21 & 3/6/21 & 0/0/21 & 0/1/21 & 2/0/21 & \textbf{14/3/21} \\
HPA & Total & 13/14/51 & 5/11/51 & 0/1/51 & 11/11/51 & 5/3/51 & \textbf{17/18/51} \\
\hline
RE & 2 & 2/0/5 & 0/0/5 & 0/0/5 & \textbf{3/0/5} & 0/0/5 & 0/0/5 \\
RE & 3 & 2/0/7 & \textbf{2/2/7} & 0/0/7 & 2/0/7 & 0/0/7 & 1/0/7 \\
RE & 4-9 & 0/0/4 & 0/0/4 & 0/0/4 & \textbf{2/0/4} & 0/0/4 & \textbf{2/0/4} \\
RE & Total & 4/0/16 & 2/2/16 & 0/0/16 & \textbf{7/0/16} & 0/0/16 & 3/0/16 \\
\hline 
DTLZ & 2 & \textbf{3/1/7} & 0/3/7 & 2/0/7 & 1/1/7 & 1/0/7 & 0/3/7 \\
DTLZ & 3 & 2/0/7 & \textbf{3/0/7} & 0/1/7 & 1/1/7 & 0/2/7 & 1/1/7 \\
DTLZ & 6 & 0/0/7 & 1/0/7 & 2/1/7 & 0/0/7 & \textbf{3/1/7} & 1/0/7 \\
DTLZ & Total & 5/1/21 & \textbf{4/3/21} & 4/2/21 & 2/2/21 & \textbf{4/3/21} & 2/4/21 \\
\hline
Minus-DTLZ & 2 & \textbf{4/0/7} & 0/0/7 & 0/0/7 & 2/0/7 & 0/0/7 & 1/1/7 \\
Minus-DTLZ & 3 & \textbf{4/0/7} & 1/1/7 & 0/0/7 & 1/0/7 & 0/0/7 & 1/1/7 \\
Minus-DTLZ & 6 & \textbf{4/1/7} & 0/0/7 & 1/0/7 & 0/0/7 & 1/1/7 & 1/1/7 \\
Minus-DTLZ & Total & \textbf{12/1/21} & 1/1/21 & 1/0/21 & 3/0/21 & 1/1/21 & 3/3/21 \\
\hline 
WFG & 2 & 2/2/9 & 0/2/9 & 1/0/9 & \textbf{5/4/9} & 0/0/9 & 1/3/9 \\
WFG & 3 & 1/1/9 & 0/0/9 & 0/0/9 & \textbf{7/1/9} & 0/0/9 & 1/1/9 \\
WFG & 6 & 0/0/9 & 0/0/9 & 0/0/9 & 1/0/9 & 1/2/9 & \textbf{7/0/9} \\
WFG & Total & 3/3/27 & 0/2/27 & 1/0/27 & \textbf{13/5/27} & 1/2/27 & 9/4/27 \\
\hline
Minus-WFG & 2 & 1/0/9 & 0/0/9 & 0/0/9 & \textbf{8/1/9} & 0/0/9 & 0/1/9 \\
Minus-WFG & 3 & 1/1/9 & \textbf{5/1/9} & 0/1/9 & 2/3/9 & 1/1/9 & 0/2/9 \\
Minus-WFG & 6 & 0/0/9 & 0/0/9 & 1/1/9 & \textbf{4/0/9} & 1/0/9 & \textbf{3/1/9} \\
Minus-WFG & Total & 2/1/27 & 5/1/27 & 1/2/27 & \textbf{14/4/27} & 2/1/27 & 3/4/27 \\
\hline
\end{tabular}}
\label{tb::igd_best_all}
\end{center}
\end{table}

\begin{table}[t]
\caption{Mean $\text{IGD}^+$ and mean evaluation time per solution. The background colors indicate the relative magnitudes of $\text{IGD}^+$ for each problem. The smallest $\text{IGD}^+$ is underlined, and $\text{IGD}^+$ that is not significantly different from the best is presented in bold.}
\centerline{\includegraphics[width=0.95\columnwidth]{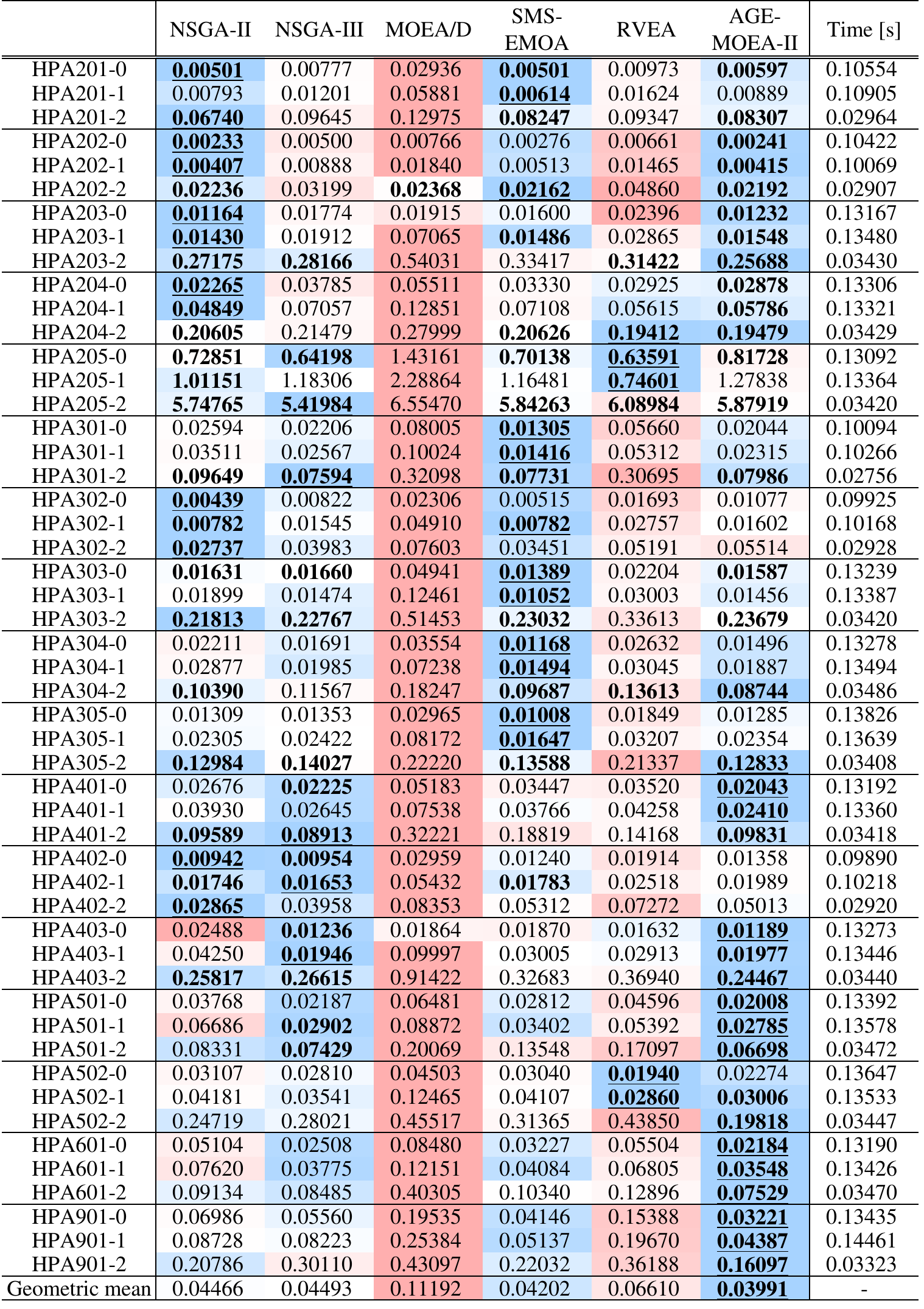}}
\label{tb::igd_detail}
\end{table}

\end{document}